\documentclass[conference]{IEEEtran}
\usepackage{times}

\usepackage[numbers]{natbib}
\usepackage{multicol}
\usepackage{graphicx}
\usepackage{amsmath}
\usepackage{amsfonts}
\usepackage{color}
\usepackage{float}
\usepackage{tabularx}
\usepackage[bookmarks=true]{hyperref}

\newcommand{\MCIGAN}{M_{\text{CIGAN}}}
\newcommand{\MIM}{M_{\text{IM}}}
\newcommand{\rpm}{\raisebox{.2ex}{$\scriptstyle\pm$}}

\let\labelindent\relax
\usepackage[inline]{enumitem}

\begin{document}

\title{Learning Robotic Manipulation through Visual Planning and Acting}



\author{
    \IEEEauthorblockN{Angelina Wang\IEEEauthorrefmark{1}, Thanard Kurutach\IEEEauthorrefmark{1}, Kara Liu\IEEEauthorrefmark{1}, Pieter Abbeel\IEEEauthorrefmark{1}, Aviv Tamar\IEEEauthorrefmark{2}}
    \IEEEauthorblockA{\IEEEauthorrefmark{1}UC Berkeley, EECS Department
    }
    \IEEEauthorblockA{\IEEEauthorrefmark{2}Technion, Department of Electrical Engineering
   }
}


\maketitle

\begin{abstract}
Planning for robotic manipulation requires reasoning about the changes a robot can affect on objects. When such interactions can be modelled analytically, as in domains with rigid objects, efficient planning algorithms exist. However, in both domestic and industrial domains, the objects of interest can be soft, or deformable, and hard to model analytically. For such cases, we posit that a data-driven modelling approach is more suitable. In recent years, progress in deep generative models has produced methods that learn to `imagine' plausible images from data. Building on the recent Causal InfoGAN generative model, in this work we learn to imagine goal-directed object manipulation directly from raw image data of self-supervised interaction of the robot with the object. After learning, given a goal observation of the system, our model can generate an imagined plan -- a sequence of images that transition the object into the desired goal. To execute the plan, we use it as a reference trajectory to track with a visual servoing controller, which we also learn from the data as an inverse dynamics model. In a simulated manipulation task, we show that separating the problem into visual planning and visual tracking control is more sample efficient and more interpretable than alternative data-driven approaches. We further demonstrate our approach on learning to imagine and execute in 3 environments, the final of which is deformable rope manipulation on a PR2 robot.

\end{abstract}

\IEEEpeerreviewmaketitle

\section{Introduction}
Many objects that we manipulate every day are deformable or non-rigid. Thus, for future robots to enter environments such as homes and hospitals, non-rigid object manipulation
will be essential. Current industrial applications such as wire threading, bin packing, and cloth folding also require such an ability. However, robotics capabilities for the general manipulation of deformable objects are still in their infancy. 

The main difficulty in planning the manipulation of deformable objects is that, in contrast with rigid objects, there is no obvious mapping from an observation of the object to a compact representation in which planning can be performed. Thus, traditional task and motion planning approaches, which require manual design of the states and transitions
in the problem, are difficult to apply~\cite{mcconachie2017interleaving,srivastava2014combined}. For example, rope manipulation involves many design aspects: should we represent the shape of rope as a finite set of small segments or as one continuous function? Should length, softness, friction, thickness, etc. be included in the model? How do we infer the system state from the robot's perception system? Different choices can be suitable for different domains, requiring substantial engineering effort. 

In recent years, several studies have proposed a data-driven, self-supervised paradigm for robotic manipulation~\cite{pinto2016supersizing,IEEEexample:poke,nair2017combining,IEEEexample:finn2017foresight,IEEEexample:ebert2017temporal}. In this approach, the robot `plays' with the object using some random manipulation policy (e.g., randomly grasping or poking an object), and collects  perceptual data about the interactions with the object. Later, machine learning is used to train a policy that performs the task directly from the perceptual inputs. By relying directly on perceptual data, these approaches overcome the modelling challenges of classical planning approaches, and scale to handle high-dimensional perceptual inputs such as raw images. 

In particular, Nair et al.~\cite{nair2017combining} learned an inverse dynamics model for rope manipulation directly from raw image data collected by randomly poking the rope. This controller was used to manipulate the rope into a given shape, whereby a human would first provide a sequence of images -- a \emph{visual plan} -- that prescribes the desired trajectory of the rope, and then the learned inverse model would compute actions that track the plan (a.k.a. visual servoing). The human demonstration in~\citet{nair2017combining} was essential -- performing high-level planning cannot be captured by a reactive inverse model. Indeed, humans' capability of planning long horizon and complex manipulations of general objects has not yet been matched by current AI technology. 

   \begin{figure}[t]
      \centering
\includegraphics[width=\columnwidth]{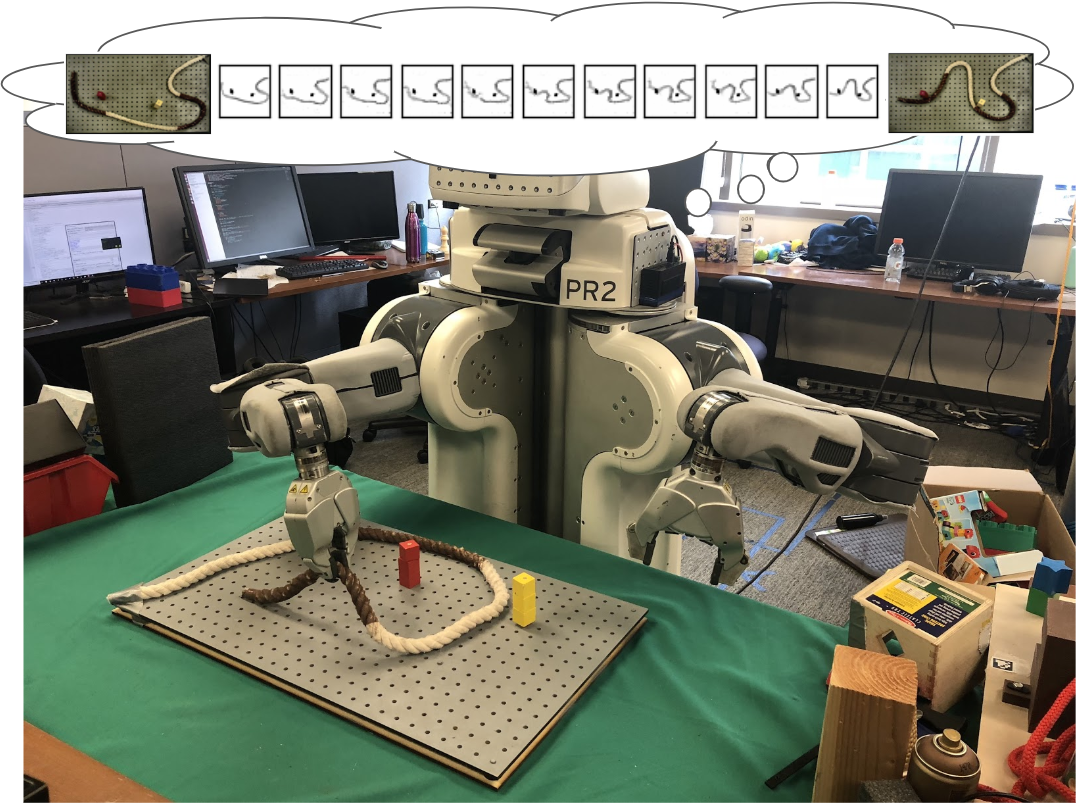}
      \caption{Visual planning and acting for rope manipulation. The PR2 robot first collects data through self-supervised random rope manipulation, and learns from this a generative model for possible visual transformations of the rope. Then, given a goal observation for the rope, we \emph{plan} a visual trajectory of a possible manipulation sequence that reaches the goal (shown on top). Finally, visual servoing is used to execute the imagined plan.}
      \label{figurelabel_intro}

   \end{figure}

In this work we take a step towards closing the gap in complex object manipulation and ask -- can we learn from self-supervised data to automatically generate the visual plan and follow it? We term this approach 
\emph{visual planning and acting (VPA)}, as depicted in Figure \ref{figurelabel_intro}. Concretely, given the current image of the system and some desired goal observation, we would like to generate a sequence of images that manipulate the object to the desired configuration, \emph{without any human guidance}, and then track this imagined plan in practice using a learned inverse model.
Such a method would not require the manual guidance of previous approaches, and would also  be \emph{safe}, as the imagined plan is visually interpretable, and can be inspected before being executed by the robot. 

However, learning visual planning from raw image data has so far been limited to very simple tasks, such as reaching or pushing rigid objects~\cite{IEEEexample:finn2017foresight,IEEEexample:ebert2017temporal}. The fundamental difficulty is that learning an accurate representation of the data requires mapping the image to an extensive feature space, while efficient planning generally requires either low dimensional state spaces or well-structured representations. Current approaches~\cite{IEEEexample:finn2017foresight,IEEEexample:ebert2017temporal} solve this tradeoff by employing very simple planning methods such as random shooting, which do not scale to more complex planning problems. 

In this work, 
we propose to learn features that \emph{are compatible} with a strong planning algorithm. At the basis of our approach is the recent Causal InfoGAN (CIGAN) model of Kurutach et al. \cite{IEEEexample:causal}. 
In CIGAN, a deep generative model is trained to predict the possible next states of the object, with a constraint that linear trajectories in the latent state of the model produce feasible observation sequences. Kurutach et al. \cite{IEEEexample:causal} used a CIGAN model for planning goal-directed trajectories simply by linearly interpolating in the latent space, and then mapping the latent trajectory to observations for generating the visual plan. 
Building on CIGAN, we propose a method for VPA, where sensory data obtained from self-supervised interaction is used to learn both a CIGAN model for visual planning and an inverse model for tracking a visual plan, as shown in Figure \ref{figurelabel_intro}. After learning, given a goal observation for the system, we first use CIGAN to imagine a sequence of images that transition the system from its current configuration towards the goal. Then, we use the imagined trajectory as a reference for tracking using the inverse model.

In this work we investigate several aspects of the VPA approach for real-world tasks. Our contributions include:
\begin{itemize}
    \item An extension of the CIGAN model to include contextual input, and imagine plans based on this context (a context can specify, e.g., obstacles in the domain), thereby addressing generalization of VPA to changes in the environment.
    \item Improvement of the planning algorithm in latent state from interpolation, as suggested in \cite{IEEEexample:causal}, to A$^*$ for planning in domains that include obstacles. 
    \item A simulation study showing that separating the control task into visual planning and visual tracking is more sample efficient than model free reinforcement learning methods that learn actions directly from images.
    \item Application of VPA to real robot rope manipulation tasks, illustrating non-trivial planning and control with deformable objects and demonstrating the interpretability of our approach.
\end{itemize}

\section{Related Work}

Deformable soft object manipulations have been attempted via classical methods such as motion planning and manipulation planning \cite{mcconachie2017interleaving,frank2011efficient,khalil2010dexterous,jimenez2012survey,saha2006motion}. These approaches require manual engineering for object models. Previous work has modeled deformable soft objects by hand-engineering representations \cite{kuniyoshi1994learning, wakamatsu2006knotting, morita2003knot, maitin2010cloth},  parametrizing the object shape \cite{miller2011parametrized}, and using finite element models \cite{hopcroft1991case}, \cite{bell2010flexible}.

Alternatively, there has been recent interest in applying learning-based approaches to robotic manipulation directly from raw image perception. 
Recent work in model-free reinforcement learning (RL) \cite{IEEEexample:dqn, IEEEexample:actorcritic, lillicrap2015continuous} learns, through trial and error, a policy mapping observations to actions that maximizes reward using deep neural networks. However, specifying reward functions for high dimensional observations such as images can be difficult \cite{IEEEexample:Arulkumaran2017survey}, and the sample efficiency of model free RL can be prohibitive in practice. Because the policy is trained to optimize a predefined reward function, it does not directly generalize to new initial and goal configurations, and requires further interactions with the system. In addition, model-free RL produces black-box policies which are hard to interpret, in contrast with more traditional planning approaches, and our visual planning method in particular, which can predict the trajectory of the robot in advance.

Learning from demonstrations (LfD) guides robots to perform complicated tasks without having to plan from scratch. Schulman et. al.  \cite{schulman2013case} and Mayer et. al. \cite{mayer2008system} learn a policy that imitates non-rigid object manipulation such as surgical suturing from expert state and action trajectories. One caveat of LfD is that it suffers when generalizing to desired trajectories that deviate from expert demonstrations. Nair et. al. \cite{nair2017combining} and Kuniyoshi \cite{kuniyoshi1994learning} only collect random interactions with the system at training time, and use the data to learn an inverse model. This inverse model is general enough to follow new expert trajectories for new tasks. In our work, we do not require expert demonstrations for new tasks, and show that visual plans can be generated directly from self-supervised data.

Other approaches that learn plannable features for control include Embed to Control (E2C)~\cite{IEEEexample:e2c} and related methods based on variational autoencoders~\cite{IEEEexample:VaST, banijamali2017robust,asai2017classical}.
Paxton et. al. \cite{IEEEexample:paxton2018visual} learn transitions and an action value function in the latent space, and use that to produce visual plans on simulated domains. To our knowledge, we present the first application of plannable features for real robot experiments. 

\section{Preliminaries and Problem Formulation}
In this section we present our problem formulation, and summarize preliminary material.
\subsection{Problem Formulation}\label{ssec:problem_formulation}
We consider a robot that interacts with the world in a self-supervised manner, and collects sensory data about its interaction. In this work, we do not consider how to collect the data, and assume that the data collection policy visits the `interesting' configurations of the system. Denote by $\mathcal{D}$ our data, in the form of $N$ trajectories of action-observation pairs, $\{o_1^i,u_1^i,...,u_{T_i-1}^i,o_{T_i}^i\}_{i\in N}$, where $u_j^i$ is the action that the robot took after observing $o_j^i$, and led to observation $o_{j+1}^i$. We assume a deterministic and fully observable system.

After we have collected the data, our goal is to solve a goal-directed planning problem: given the current observation of the system $o_{start}$ and an observation of a desired goal configuration $o_{goal}$, we want to compute an action selection policy that transitions the system from start to goal.

To solve the problem above, in this work we focus on an approach we term Visual Planning and Acting (VPA). The idea is to decompose the solution into two steps:
(1) Visual planning -- learning from the data how to \emph{imagine} a goal-directed trajectory of observations that transition the system from start to goal, and (2) Acting -- using an inverse model learned from the data on how to take actions that make the system follow the imagined plan. 

\subsection{Visual Planning with Causal InfoGAN}
Kurutach et al.~\cite{IEEEexample:causal} describe a method for visual planning based on the CIGAN generative model. Before describing CIGAN, we first summarize two ideas that it builds on -- GAN and InfoGAN.

\subsubsection{GAN and InfoGAN}
GANs \cite{IEEEexample:gan} are deep generative models that learn to generate samples similar to the data distribution $P_{data}$ by feeding in a random vector $z \sim \mathcal{N}(0,I)$ into a deep neural network generator $G$. 
A discriminator neural network $D$ tries to tell apart generated samples from real samples, and the GAN training objective is given by the minimax game:
$
\min_G \max_D V(G, D)=\\
\min_G \max_D \mathbb{E}_{o \sim P_{data}}
[\log D(o)]+\mathbb{E}_{z}[\log (1 - D(G(z)))].
$
The vector $z$ can be interpreted as a latent representation for the generated observation $o=G(z)$.
InfoGAN~\cite{IEEEexample:info} is a method for adding structure to the latent representation. In InfoGAN, the representation is separated into a `noise' component $z$ and a structured component $s$. The loss function is modified to maximize the mutual information between $s$ and the generated observation $o=G(z,s)$, which intuitively induces $s$ to capture salient properties of the observation.
Let $x,y$ be some random variables. Denoting $H(x)=\mathbb{E}_x [-\log (P(x))]$ as the entropy of $x$, the mutual information between $x$ and $y$ is defined as $I(x;y)=H(x)-H(x|y)=H(y)-H(y|x)$. The InfoGAN loss function is:
\begin{equation}\label{eq:infogan_loss}
\min_G \max_D V(G, D)-\lambda I(s;G(z,s)).
\end{equation}
To optimize this loss in practice, a variational lower bound was proposed in \cite{IEEEexample:info}. Let $Q(s|o)$ denote an auxiliary distribution that approximates the posterior $P(s|o)$. Then the lower bound $I(s;G(z,s)) \ge \mathbb{E}_{s\sim P(s), o\sim G(z,s)} [\log Q(s|o)]+H(s)$ can be plugged in \eqref{eq:infogan_loss} and optimized using the reparametrization trick~\cite{IEEEexample:info}. Intuitively, the $Q$ function can be understood as a classifier that encodes an observation into its latent representation.

\subsubsection{Causal InfoGAN and Plan Generation}\label{ssec:background_CIGAN}
CIGAN~\cite{IEEEexample:causal} is an extension of InfoGAN for observations from a dynamical system. Consider data that contains trajectories of observations, similar to $\mathcal{D}$ described above in Section \ref{ssec:problem_formulation}. The CIGAN model learns to generate a pair of \emph{sequential} observations $(o,o')$ that are similar to sequential observations in the data, thereby learning a notion of \emph{causality} in the data. The CIGAN generator input is a pair of latent representations $s,s',$ and a noise vector $z$: $o,o' = G(s,s',z)$, where similarly to InfoGAN, the intuition is to learn a transition in the latent space $s\to s'$ that captures salient properties of the observation transition $o \to o'$. 

In \cite{IEEEexample:causal}, the latent state distributions were $P(s)= \mathcal{N}(0,I), P(s'|s) = \mathcal{N}(s,\sigma(s))$. That is, the next state $s'$ was modelled as a local perturbation of the first state $s$, where the magnitude of the perturbation $\sigma(s)$ was a learned neural network. The motivation for such dynamics was to structure the latent space to be compatible with a planning algorithm, as described below. The CIGAN loss is similar to InfoGAN, with the additional learning of dynamics in latent space $P(s'|s)$, and the mutual information between the \emph{pairs} of latent states and observations:
\begin{equation}\label{eq:CIGAN_loss}
\begin{split}
\min_{P(s'|s), G} \max_D V(G, D)-\lambda I(s,s';o,o'),\\
\text{s.t. } o,o'\sim G(z,s,s'),    s\sim P(s),      s'\sim P(s'|s)
\end{split}
\end{equation}
To optimize \eqref{eq:CIGAN_loss}, an InfoGAN lower bound was used, introducing an auxiliary distribution $Q(s,s'|o,o')$ to approximate $P(s,s'|o,o')$. Kurutach et al. proposed to use a disentangled approximation $Q(s,s'|o,o')=Q(s|o)Q(s'|o')$.

\begin{figure}[]
      \centering
\includegraphics[width=\columnwidth]{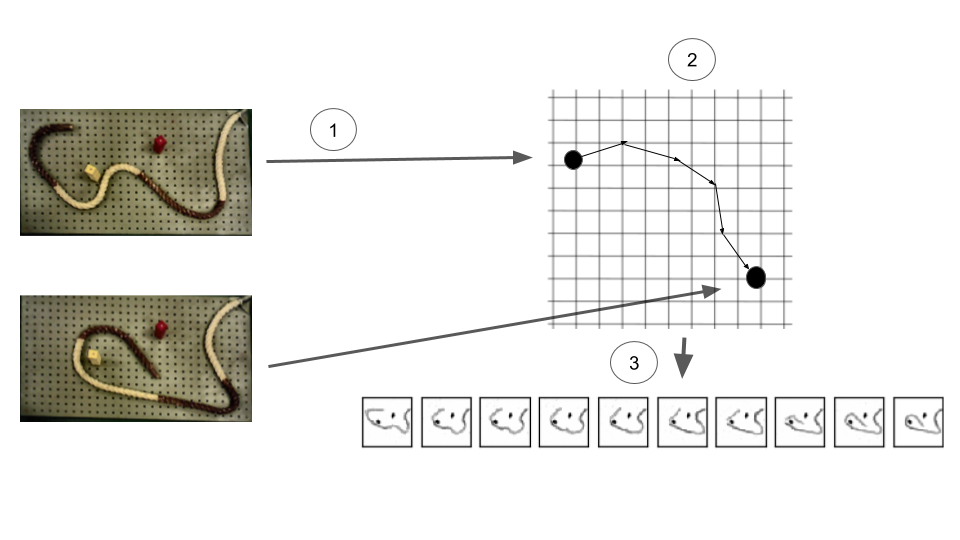}
      \caption{Illustration of how the CIGAN model generates a plan. First, start and goal images are encoded to their latent representations (denoted here as points in the plane). Second, search is used to find a sequence of points in the latent space that connect the start to the goal, while obeying the latent space dynamics. Here we illustrate the result of A* search. Third, the plan in latent space is decoded into a sequence of images using the generator, resulting in a visual plan.}
      \label{fig:ciganplan}

   \end{figure}

A CIGAN model trained on the data $\mathcal{D}$ can be used for visual planning according to the following scheme~\cite{IEEEexample:causal} and visually represented in Figure \ref{fig:ciganplan}:
\begin{enumerate}
\item Encoding: given a pair, $o_{start}, o_{goal}$, find the corresponding $s_{start}, s_{goal}$.
\item Planning: in the latent space, find a feasible trajectory: $s_{start}, s_1,...,s_{m}, s_{goal}$
\item Decoding: from the latent trajectory generate a feasible observation trajectory $o_{start}, o_1,...,o_{m}, o_{goal}$.
\end{enumerate}

For encoding, Kurutach et al.~\cite{IEEEexample:causal} used an optimization based approach, searching for a latent vector that minimizes the absolute pixel difference with the desired observation.
For the planning, the key idea in~\cite{IEEEexample:causal} is that due to the local transition structure of $P(s'|s)$, linear interpolation between $s_{start}$ and $s_{goal}$ results in a feasible plan. In this sense, CIGAN learns a representation that is \emph{compatible} with the planning algorithm. For decoding, the CIGAN generator can be used to sequentially produce pairs of observations from the trajectory.

\subsection{Learning Inverse Dynamics Models}
An inverse model $\MIM$ maps a pair of sequential observations $o,o'$ to an action that generated them $u = \MIM(o,o')$. This can be cast as a supervised learning problem, by regressing from $o_t,o_{t+1}$ in the data to $u_t$. Here, we follow the approach of Nair et al.~\cite{nair2017combining}, which learned inverse models from image observations using deep convolutional neural networks.
Given a reference trajectory in image space $o_1^{ref},\dots,o_t^{ref}$, an inverse model can act as a tracking controller (a.k.a. visual servoing~\cite{espiau1992new}) by taking the action $\MIM(o_t,o_{t+1}^{ref})$ at time $t$.

\section{Visual Planning and Acting}\label{sec:method}
In this section we present our approach for solving the goal directed planning problem of Section \ref{ssec:problem_formulation}, which we term Visual Planning and Acting (VPA). 

Our approach is model-based, where we first use the data $\mathcal{D}$ to learn both a CIGAN model $\MCIGAN$ and an inverse dynamics model $\MIM$. For any two start and goal observations $o_{start},o_{goal}$, the CIGAN model $\MCIGAN$ can generate a visual plan that transitions the system from start to goal, $o_{start}, o_1, \dots, o_k, o_{goal}$. Since the CIGAN model is trained to generate feasible pairs of observations (cf. Section \ref{ssec:background_CIGAN}), the plan generated by a well-trained CIGAN model will be feasible, in the sense that the robot can actually execute it.

Our VPA method for solving the goal directed planning problem is a combination of planning and replanning using the CIGAN model $\MCIGAN$, and trajectory tracking using the inverse model $\MIM$. The VPA algorithm is given as follows:
\begin{enumerate}
\item \textbf{Plan:} given a pair, $o_{start}, o_{goal}$, use the CIGAN model $\MCIGAN$ to generate a planned sequence of observations $o_{start}, o_1,...,o_{m}, o_{goal}$. 
\item \textbf{Act:} If the length of the plan $m$ is zero, take an action $u$ to reach the goal $u = \MIM(o_{start},o_{goal})$, then stop. Else: 
\item Take an action $u$ to reach the first observation in the plan $u = \MIM(o_{start},o_{1})$ and take a new observation of the current system state $o_{new}$.
\item \textbf{Replan:} update $o_{start}$ to be the current observation $o_{new}$, and go back to step $1$.
\end{enumerate}

VPA effectively uses the inverse model as a feedback controller to follow the imagined CIGAN plan. In practice, we found that the advantage of replanning in our tasks was not significant, and chose to omit this step for faster execution times. That is, instead of replanning from the current observation we simply advanced on the original plan by removing the first observation $o_1$. However, other tasks may benefit from full replanning.

We emphasize that while VPA uses planning, it builds on CIGAN, which is completely data-driven, and does not require manually engineering a planning model. The only data required for this is images taken from self-supervised manipulation of an object. Nevertheless, our method enjoys the \textbf{interpretability} of model based methods -- at every step of our algorithm we have a visual plan of the proposed manipulation. We found that this allows us to reliably evaluate the performance of VPA before performing any robot experiment, significantly reducing time and effort as well as unpredictability in the robot's actions.  
We also remark that separating decision making into a high-level trajectory computation step and a low-level action execution is standard in motion planning~\cite{lavalle2006planning}, and has been explored in several recent studies on robotic manipulation~\cite{sung2018robobarista,thomas2018learning}. Here, in comparison, the trajectories are in \emph{image space}, and hence can capture complex object features such as deformations and change in appearance. Another benefit of separating observations and actions is the possibility of collecting different data for training $\MCIGAN$ and $\MIM$. For example, in rope manipulation, the properties of the rope are largely independent of the robot manipulating it. Thus, we can collect a robot-independent data for training $\MCIGAN$ with several different robots, or even a human, as we did in our experiments, and then collect a robot-specific data set for training $\MIM$ for a particular robot.\footnote{In principle, the action can be subsumed in the observation for training a CIGAN model that can plan actions. Due to the benefits mentioned above, we opted for computing actions independently using an inverse model.}

These properties makes our approach suitable for deformable object manipulation, as we demonstrate in our experiments. However, 
in order to get VPA to work well in practice, we needed to make several fundamental changes to the CIGAN method, as we describe next. We also describe several technical modifications in Appendix \ref{sec:appendix}.

\subsection{Context Conditional CIGAN Model}
The CIGAN model in~\cite{IEEEexample:causal} generates observations that are, by definition, similar to the training data. That precludes any \emph{generalization} to problem parameters that are different than those seen during training. However, in practical settings, we would like to generalize our knowledge to change in the environment. For example, in an environment with obstacles, one would like to learn a model that can generalize to different obstacle configurations. 

Here we approach this problem by adding to the CIGAN model a \emph{context} input. We assume that 
the domain can be decomposed into a manipulatable, movable object (e.g., rope), and components which are fixed during manipulation (e.g., obstacles). 
We propose a modification of the CIGAN architecture that takes as input an observation of the fixed components as a \emph{context vector}. We term this model a Context Conditional CIGAN (C$^3$IGAN). By training on a variety of context vectors, we should hope that the model generalizes to novel contexts.

\begin{figure}[thpb]
      \centering
      \parbox{3.3in}{\includegraphics[scale=.37]{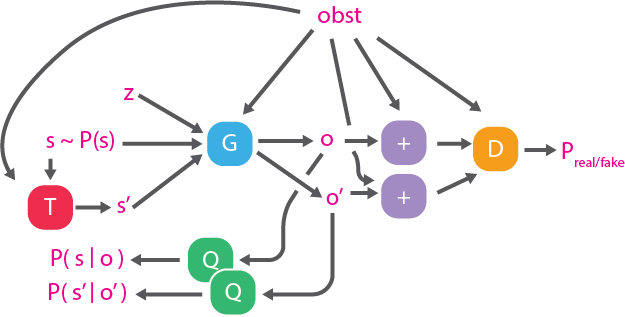}
}
      \caption{C$^3$IGAN Model Architecture}
      \label{fig:c3igan}
   \end{figure}

In C$^3$IGAN, as shown in Figure \ref{fig:c3igan}, the generator takes in as input $z, s, s', c$, where the context $c$ represents an image of the fixed components in the domain, so in our case, the obstacles (obst). The generated observations are added (pixel-wise) to $c$ before they are passed onto the discriminator. In this way the generator is trained to generate only the movable part in the scene.
Thus, the generator is now only in charge of generating the images of the rope, and not the obstacles. By relieving the generator of the responsibility of generating a fixed backdrop that is fixed throughout a trajectory, it can focus on the nuances of the object whose movement we actually want to control. 
The generator in this model is also able to generalize to new obstacle configurations not seen during training. This model has the potential to extend to other applications where there is a fixed background to interact with, such as a maze or other physical barriers.
\subsection{A$^{*}$ Planning in Latent Space}

As discussed in Section \ref{ssec:background_CIGAN}, Kurutach et al.~\cite{IEEEexample:causal} used a simple linear interpolation in latent space as the planning computation, based on the insight that the latent state transitions in CIGAN are Gaussian perturbations, guaranteeing that a small step in any direction in the latent space should result in a feasible transition. In our experiments on rope manipulation, we found that this method did not produce realistic enough of plans, especially in the presence of obstacles, which it tended to go through rather than around. 
We attribute this to the fact that the actual covariance matrix of the Gaussian probability may be asymmetric such that some directions have extremely low likelihood.

To account for this, we learned a more expressive transition model, $P(s'|s)\sim N(s+\delta(s), \sigma(s))$, where both $\delta$ and $\sigma$ are neural networks. Shifting the mean allows the transition to prefer some direction than the others.
We added a loss on the magnitude of $\delta$ and $\sigma$ in order to induce a local transition structure, which complies with the A$^*$ heuristic and allows us to perform such planning at test time.

With this new transition model, we propose a different approach for planning in the latent space, which combines sampling and A$^*$ directed search. Given any state $s$, we can sample $N$ possible next states $s'$ from the probability $P(s'|s)$. Thus, we can recursively build a sampled connectivity graph of the possible transitions in latent space. Potentially, we can search this graph for a trajectory that reaches from $s_{start}$ to a state close to $s_{goal}$. However, in practice, the latent space dimensions are too large to perform a naive search in reasonable time. To solve this problem, similar to Kurutach et al.~\cite{IEEEexample:causal}, we leverage the \emph{structure} of the latent space, and in particular, the local connectivity structure enforced by the Gaussian transition model. We propose to use the directed search algorithm A$^*$ with the Euclidean distance as a heuristic function, utilizing the fact that with local transitions, the Euclidean heuristic is admissible~\cite{IEEEexample:ai}. More detail is provided in Appendix Section \ref{app:astar}.

To improve the precision of our plans, which in practice are imperfect, we supplemented the sampling method by pruning unfeasible transitions, using a separately trained classifier,
trained on positive examples of perturbed real rope images to bring the distribution closer to that of the generated data, and negative examples of generated rope images. The reason for using the separate classifier for this task rather than the CIGAN's discriminator, is that the discriminator overfit to classifying all generated images as fake, and was unable to distinguish between fake and good transitions, and fake and bad transitions.
For each sampled pair of states $s,s'$, we generate a corresponding observation transition $o,o'$ from the CIGAN generator, and if the classification score for this pair is lower than a threshold, we prune the transition $s\to s'$ from the connectivity graph.

\section{Experiments}

We designed our experiments to address the following questions:
\begin{enumerate}
    \item Can our method generate non-trivial visual plans, and is the fidelity of these visual plans high enough to be combined with an inverse model for plan execution?
    \item How does VPA compare to alternative methods like batch RL or running the inverse model without a plan?
    
    \item Can VPA leave the simulation and work on a real robot?
\end{enumerate}

We demonstrate our method on three domains. The first is a two-block world in Mujoco~\cite{todorov2012mujoco}. In this domain, we perform a comparison with batch off-policy RL -- an alternative method for learning a control policy from data.
The second domain contains a movable block with a static obstacle. In this domain, we show the need for planning when the inverse model fails to navigate around the obstacle, while VPA learns to do so.
Finally, we deploy the algorithm on a PR2 robot to manipulate a deformable rope around obstacles. Within real world rope manipulation, we explore two similar variations of the domain: one with static obstacles in which we compare our method to that of Nair et al.~\cite{nair2017combining}, and the other with dynamic obstacles in which we demonstrate the potential of generalizing to variations in the environment using C$^3$IGAN.

\subsection{Two-Block Domain}
   
In this domain, the task is to move two rigid blocks on a table to some goal location. Possible actions are moving a block by some small offset in any direction. The table is $1.5$ units on each side, and we consider the task a success if the L2 distance between the final and goal states is below $0.5$. 
We collect data by randomly applying actions in the domain, and structure our data to contain 30k observation transitions, where only 2K transitions also have action labels. This corresponds to a setting where collecting possible observation transitions is easier than collecting real robot actions, as described above, and also demonstrated in our real robot experiment.
 
\begin{figure}[thpb]
  \centering
  \parbox{3.3in}{\includegraphics[scale=.35]{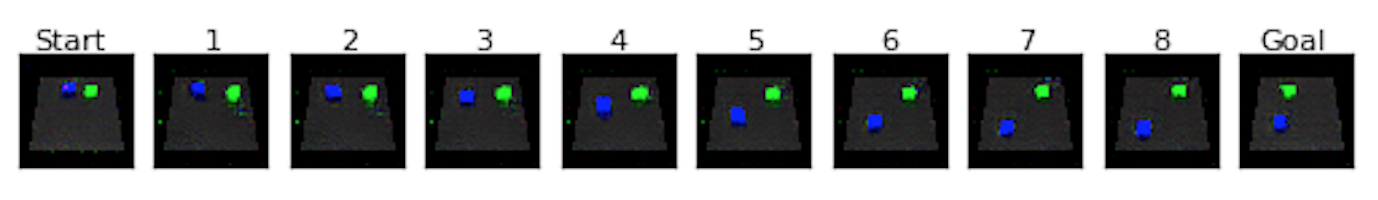}
  
  \includegraphics[scale=.35]{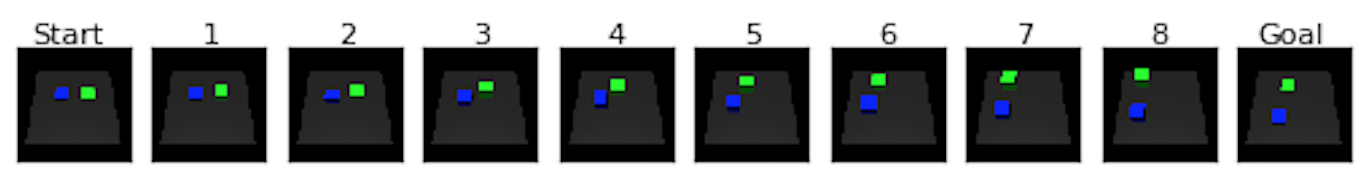}
}
  \caption{Top image: Visual Planning step - imagined plan by Causal InfoGAN. Start and Goal image are both $o_{closest}$ to the actual $o_{start}$ and $o_{goal}$, which are shown right below them. Bottom image: Execution step - images showing the actual successful results of running entire VPA pipeline on Mujoco}
  \label{fig:block_plan}
\end{figure}

For VPA, we train a CIGAN model on the full dataset, and an inverse model on the action-labeled data. We use linear interpolation for planning, as in~\cite{IEEEexample:causal}, as this domain is simple enough to not require the more complex A$^*$. In Figure \ref{fig:block_plan}, we show a sample plan generated by CIGAN, and the corresponding trajectory executed by VPA. It can be seen that the initial plan is visually interpretable, and resembles the actual trajectory that was executed. Quantitatively, we evaluated VPA on 50 random initial and goal configurations that were not in the data, as shown in Table \ref{tab:block_eval}. 

We compare VPA with an alternative data-driven approach based on model-free batch RL\footnote{For image observations, the state of the art in RL is model free~\cite{mnih2015human,lillicrap2015continuous,kalashnikov2018qt}, while recent model based approaches are limited to lower dimensional observations~\cite{kurutach2018model}. Therefore, we did not consider model based RL in our comparison.}, namely, fitted Q-iteration~\cite{riedmiller2005neural} (equivalent to a single epoch of DQN~\cite{mnih2015human} with the data as the replay buffer). 
RL requires action labels, so we trained with only the action-labeled part of the data. Since actions are continuous, we used random sampling to find the maximal Q value in the Bellman backup, similar to~\cite{kalashnikov2018qt}. For the state space, we embedded the images into a latent space using a variational autoencoder, \emph{trained on all the data}, and the reward was based on distance in latent space, as recently suggested in~\cite{nair2018visual}. Since RL is not expected to generalize, we \emph{retrained the Q network on all the data for each goal in the evaluation}.
This is a strong baseline, that makes use of both the action-labeled and unlabeled data, incorporates several recent techniques for image-based RL, and our evaluation forgives the limitations of RL in generalizing to different goals. 

However, as stated earlier, RL is known to have difficulties with large state spaces (image), reward specification, and sample efficiency. To demonstrate this, we also run RL with several \emph{artificial benefits}: (1) simple state space -- true positions of the blocks, (2) true reward -- based on real distance to target, and (3) significantly more data -- 30k action-labeled samples.

Our results, reported in Table \ref{tab:block_eval} show that, surprisingly, VPA significantly outperforms RL even with the artificial benefits. Only with all benefits added, does RL compare well with VPA. We attribute these results to the fact that in this domain, decomposing control to trajectory planning and tracking control is natural, and VPA exploits this structure. Indeed, the common failure case for RL is pushing the blocks off the table due to an inaccurate Q function. Since VPA never imagines plans where blocks go off the table, our method was resilient to such failures.

\begin{table}[h!]
\caption{The average final L2 distance to goal and the success rate of moving two blocks to be within 0.5 radius to the goal when executed on 50 new tasks.}
\centering
\begin{tabular}{ |c|c|c| }
 \hline
 Method & L2 distance & Success Rate\\
 \hline
 VPA (2k) & \textbf{0.335 \rpm 0.121} & \textbf{90\%} \\  
 \hline
 Batch RL (positions, real $r$, 2k) & 0.657 \rpm 0.701 & 76\% \\
 Batch RL (positions, real $r$, 30k) & 0.675 \rpm 0.739 & 74\% \\
 \hline
 Batch RL (image, real $r$, 2k) & 1.172  \rpm 0.991 & 16\% \\   
 Batch RL (image, real $r$, 30k) & 1.186 \rpm 0.940 & 42\% \\
 \hline
 Batch RL (image, embedded $r$, 2k) & 1.346 \rpm 0.891 & 14\%\\
 Batch RL (image, embedded $r$, 30k) & 1.445 \rpm 1.096 & 18\%\\
 \hline
\end{tabular}
\label{tab:block_eval}
\end{table}

\subsection{Block-Wall Domain}
To further motivate the need for planning, we investigate the efficacy of our model on another simulated domain, now with planning more intuitively necessary to complete the task. In this domain, the agent has to manipulate a green block around a red vertical obstacle. We perform the same VPA method as before, on a new test set of 20 start/goal image pairs.

We compare two variations of our method against the baseline of using only an inverse model, as used in Nair et al.~\cite{nair2017combining}. 
The first method executes the single best hand-selected plan from many generated by the CIGAN. There is some variability in the quality of generated plans due to the random noise, and some are better than others, so we choose only the best one to execute.
The second method autoselects a generated plan to execute. This plan is selected using a combination of a classifier trained on the dataset and an object detector trained on a simple shape dataset. We describe this more in Appendix \ref{sec:autoselect}.

\begin{figure*}
  \centering
  \parbox{9.9in}{\includegraphics[scale=.6]{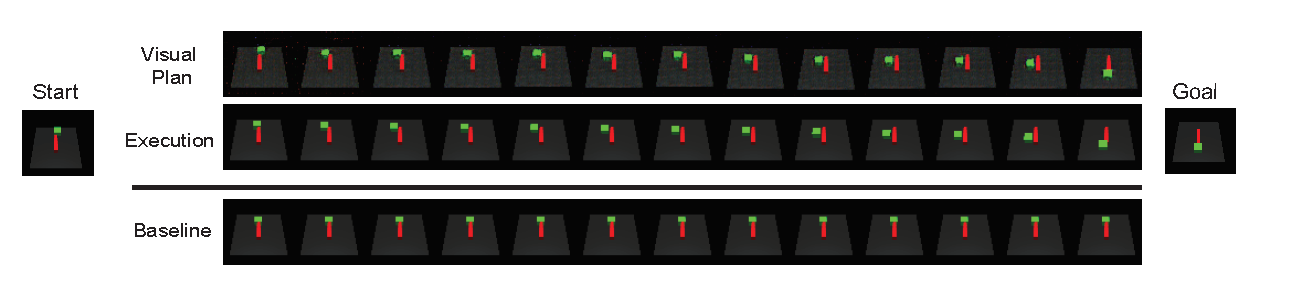}
}
  \caption{Comparison between VPA and an inverse model baseline. The baseline attempts to directly apply the inverse model on the goal, while our method employs the plan generated by CIGAN, shown at top, to navigate from start to goal. Without a plan, the baseline blindly attempts to move the block downwards without accounting for the obstacle in the way. Our model, on the other hand, plans to go around the obstacle, resulting in a successful trajectory.}
  \label{fig:block_wall}
\end{figure*}

Our results in Table \ref{tab:block_eval_wall} show that planning with VPA significantly improves upon the inverse model baseline, for both plan selection methods. 
In Figure \ref{fig:block_wall}, we show an example where planning is necessary, and the baseline is unable to execute the task while our method is successful.

\begin{table}[h!]
\caption{The average final L2 distance to goal and the success rate to move one block in the block-wall domain to be within 0.5 radius to the goal.}
\centering
\begin{tabular}{ |c|c|c| }
 \hline
 Method & L2 distance & Success Rate\\
 \hline
  Baseline & 0.459 \rpm 0.433 & 45\% \\   
 \hline
 VPA (hand-selected) &  \textbf{0.083 \rpm 0.192} & \textbf{95}\% \\
 \hline
  VPA (autoselected plan) & 0.131 \rpm 0.242 & 90\%\\
 \hline
\end{tabular}
\label{tab:block_eval_wall}
\end{table}

\subsection{Real Robot Rope Manipulation Domain}

Finally, we bring our method out of simulation and into the real world by conducting experiments with a PR2 robot manipulating a flexible rope that is fixed on one end and can move between two obstacles. This domain is inspired by wire threading -- an important industrial task that requires complex planning of rigid and non-rigid object interaction.

\subsubsection{Static Obstacles}
We begin our investigation by comparing our planning based method to a baseline of only using an inverse model without planning, as in the previous block and wall domain. We designed a rope manipulation environment similar to~\cite{nair2017combining}, but which also contains fixed obstacles which the rope cannot move through.

\begin{figure}[thpb]
  \centering
  \parbox{3.3in}{\includegraphics[scale=.23]{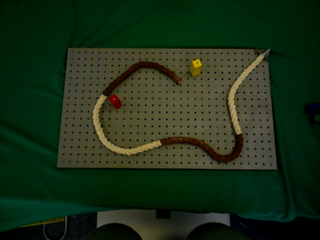}
  \hspace{0.2cm}
  \includegraphics[scale=.23]{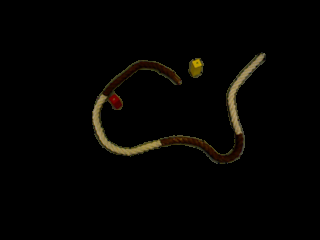}
  \hspace{0.2cm}
  \includegraphics[scale=0.87]{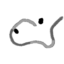}
}
  \caption{Preprocessing procedure. Image 1: original photo as seen by the PR2, Image 2: background removed from image, Image 3: Preprocessed to have one color channel}
  \label{fig:preprocess_rope}
\end{figure}

For data collection, we followed the approach in~\cite{nair2017combining} for generating random pokes of the rope, and collected 2k samples for observations and actions. To increase the size of our dataset, we collected 10k additional observation samples by manually manipulating the rope (which is much faster to collect). Note, however, that due to the obstacles, our problem is \emph{much} more difficult than in~\cite{nair2017combining}. 
The images are preprocessed to have one color channel, as demonstrated in Figure \ref{fig:preprocess_rope}.

With the additional constraint of obstacles, we conjecture that the inverse model, which is essentially reactive in its computation, will not suffice to plan movements that involve these obstacles. 
In contrast, a well trained CIGAN model should generate plans that manipulate the rope around such obstacles. For the inverse model, $\MIM$, we used a CNN similar to~\cite{nair2017combining}, outputting a 4-dimensional action made up of the $x$ and $y$ coordinates of the rope grab and drop location. We found that feeding in to $\MIM$ the observation difference $o'-o$ resulted in a more robust controller that generalized well to the generated image sequences. 
As described in Section \ref{sec:method}, to reduce computation time we did not replan in the VPA algorithm, and simply ran $\MIM$ multiple times for each generated observation, to result in a longer, but more stable execution.

\begin{figure*}
  \centering
  \parbox{9.9in}{\includegraphics[scale=.5]{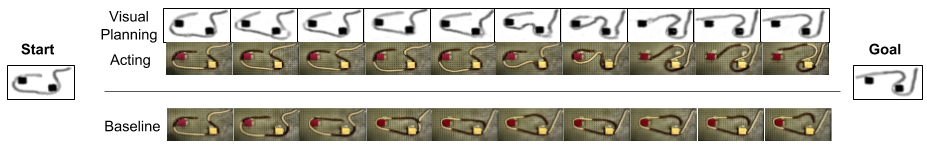}
}
  \caption{Comparison between VPA and an inverse model baseline. The CIGAN-generated plan is presented in grayscale, with the results after the PR2 robot successfully runs iterations of the $\MIM$ by tracking the plan is shown below it. The baseline of only the inverse model, $\MIM$ is shown below. Note that VPA plans to go \emph{around} the obstacle, leading to a successful plan execution, while the inverse model is not capable of such planning due to its reactive, short-term nature, and therefore cannot complete the task.}
  \label{fig:robobase}
\end{figure*}

\begin{figure}[]
  \centering
  \parbox{2.5in}{\includegraphics[scale=.45]{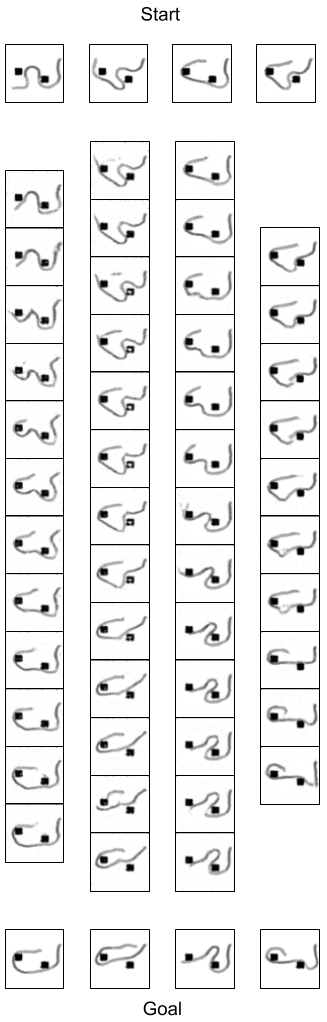}
}
  \caption{Demonstration of CIGAN plans for the rope domain with fixed obstacles. The top row shows the start states and the bottom row shows the goal states of 4 different problems given to the CIGAN. The middle 4 columns show sample generated plans. Note the realistic transitions of the rope around the obstacles, which obey physical properties of the rope such as being stretched when pulled from the end.}
  \label{fig:static_plans}
\end{figure}

In Figure \ref{fig:robobase}, we demonstrate a setting where nontrivial planning is required to solve the task -- going from start to goal requires traveling with the rope around the obstacle.
It can be seen that our VPA method \emph{plans} to go around the obstacle, which makes it feasible to solve the task by following the plan with the inverse model $\MIM$. Just using $\MIM$ on the goal image, however, does not result in traveling around the obstacle, which leads to a failure in execution. This result demonstrates the inadequacy of purely reactive methods, such as inverse models, for acting in complex domains with more constraints. We further evaluate the planning capability of our method in Figure \ref{fig:static_plans}, where we demonstrate realistic plans of rope manipulation that obey the physical properties of the rope and obstacles.

\subsubsection{Dynamic Obstacles}
In this section we demonstrate the potential of C$^3$IGAN in generalizing to unseen environments. To this end, we modified the rope manipulation domain to include \emph{dynamic}\footnote{By which we mean that the position of the obstacles change between trajectories, and not during}, smaller obstacles, which were intermittently moved (manually) while collecting the training data. These changes render this variation harder than the previous one for our vision-based planning method. Our hope is that our model can imagine, plan, and execute rope manipulation in domains with obstacle configurations that were not explicitly seen during training.

\begin{figure}
      \centering
      \parbox{3.6in}{\includegraphics[scale=.6]{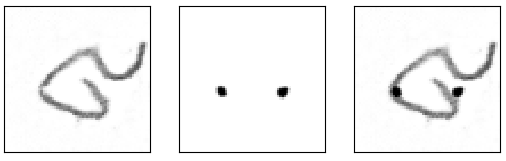}
}
      \caption{Visualization of $C^3IGAN$ results. The model generates an image conditioned on the obstacles, and then performs a pixelwise addition of the left and middle image. Left: generated rope, Middle: obstacles conditioned on, Right: pixelwise addition of images.}
      \label{obst_conditioned}
   \end{figure}

\begin{figure}[thpb]
      \centering
      \parbox{28.3in}{
    \includegraphics[width=0.48\textwidth]{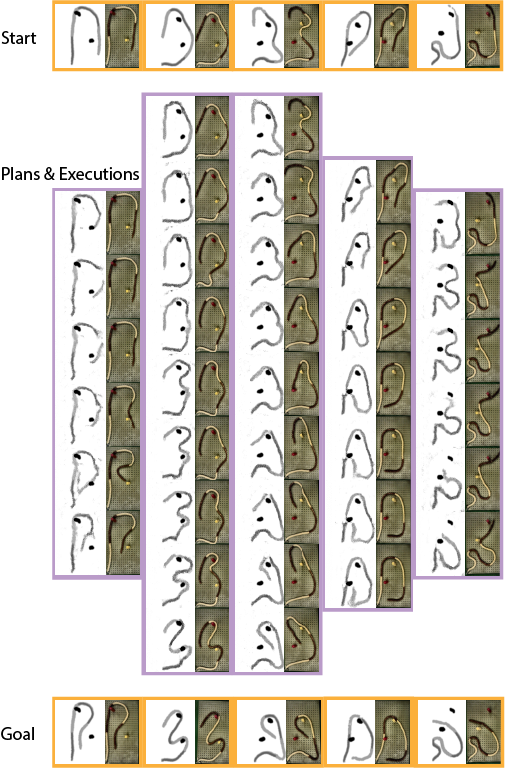}
}
      \caption{5 examples of VPA executed on the rope domain. The left 4 are successful runs, and the right 1 is where a plan is generated to reach the goal, but the action policy is not strong enough to carry it out. Looking at one column at a time, the top image is the start state and the bottom is the goal state. In the middle, the grayscale images are the visualized plan, and the colored images are the actual results of the rope when we run the inverse model to have the PR2 take actions.}
      \label{fig:rope_results}
   \end{figure}

In training our C$^3$IGAN model for this domain, the context is an image of the obstacles without any rope, as show in Figure \ref{obst_conditioned}. 
As the figure demonstrates, the obstacle embedding is successfully used by the C$^3$IGAN model to generate images that realistically capture the interaction between the rope and obstacles, such as that the rope has to wind around the obstacle, rather than moving through it.
In Figure \ref{fig:rope_results} we demonstrate results for VPA on this domain. We present both the imagined C$^3$IGAN plans, and the resulting trajectories when running VPA on the robot. 

In terms of success rate, we qualitatively inspected the plans and found that approximately $15$\% were visually plausible, which means they realistically follow what the robot is allowed to do. The most common failure cases were inaccurate encoding, leading to a misspecified goal image, or the rope breaking and reconnecting somewhere else during the trajectory. 
We believe that more data would significantly improve these results, as our results for the static obstacles, which was an easier domain and trained using the same amount of data, were indeed significantly better.
From the visually correct plans, the inverse model was able to successfully execute $20$\%. This is somewhat worse than the results of Nair et. al.~\cite{nair2017combining}, which we attribute to the order of magnitude smaller dataset we used, and our additional obstacles. We emphasize that even though our success rates are not high, most failure cases can be caught \emph{by visual inspection}, without running the robot, since our method is readily interpretable. Such interpretability has additional important implications to safety.. 
Thus, while further investigation is required to improve the quality of VPA, we see these results as a proof of concept for a promising robotic manipulation paradigm.

\section{Discussion and Conclusion}
We proposed a new data-driven paradigm for robot manipulation by learning a model for planning in image space, and using the imagined plan as a reference for a visual tracking controller. Our method is interpretable, and we showed that it can outperform model-free RL approaches in simulation. We have shown promising results using a PR2 robot in learning to manipulate a deformable rope among obstacles.

Recently, data-driven approaches to robotic manipulation have increased in popularity, and are widely considered to be essential for bringing robots into human centered environments. Most of the work so far has focused on model free RL, which, although capable of learning complex policies, results in opaque controllers that are only verifiable by physical evaluation on the robot. This has raised many issues of safety and interpretability~\cite{amodei2016concrete}, and at present there is no principled method for explainable RL.
We believe that the most important aspect of our approach is its visually interpretable nature. As we have shown, the imagined plans are easy to understand, and when supervised by a human operator, unsafe or undesired plans can be intercepted before the physical execution of the task.

In future work we will investigate how to improve the capabilities of VPA. Directions we focus on include: (1)  improving the visual tracking controller using RL or model-predictive control; (2) investigating active data collection policies that focus on the most `interesting' parts of the state space; (3) extending our approach to domains where the objects are not static between manipulation actions using memory based models; and (4) investigating realistic applications such as packing soft and rigid objects into a box or organizing objects in a cabinet.

\clearpage
\newpage

\bibliographystyle{abbrvnat}
\bibliography{references}

\clearpage
\newpage

\begin{appendix}
\label{sec:appendix} 
\subsection{Training Details}
 
In the CIGAN training, we used the same architecture for all domains from \cite{IEEEexample:causal}, but doubled the number of filters at each layer of the generator, discriminator, and posterior model ($Q$) when we moved domains from the blocks to the rope, because models with more parameters were needed to capture the nuances for the more complex domain. Additionally, we used GAN training techniques of spectral normalization~\cite{miyato2018spectral}, instance noise~\cite{instancenoise}, and label smoothing~\cite{gantraining}. We also got rid of the random noise, $z$, from the final models that were run on the rope, because we found that the added noise would end up influencing the transitions too much, and they were not actually needed to provide diversity to the generations, as the latent codes were already sufficient for that. Both the discriminator and generator were trained in unison with the Adam optimizer \cite{adam2014}.

The latent space of our model had a dimension size of 10. In order to improve the latent space learned by the model, we gradually increase the information weight $\lambda$ by .01 starting from .01 every 5 epochs. We already knew that by increasing the weight on the information term we could improve the learned representations and encode more meaning about transitions into them. However, if we increased the weight on this term any higher than the .1 we used on the blocks, the generation quality and fidelity became too poor to be able to follow and execute a control policy on. However, once the generation quality is at a certain level, there would be no reason for the generator to start generating images of worse quality, even if its loss function did change (by a small amount, so as not to disrupt the fickle stability of a GAN). Thus, gradually increasing the information weight had an improvement on both the quality of images generated by the CIGAN, as well as the causal meaning captured by the latent space. This was shown by the improvement in plans generated from this model, as they went through obstacles far less, and had better image quality.

\subsection{Plan Autoselection in Block-Wall Domain}\label{sec:autoselect}
We want to be able to autoselect the plans that seem feasible and allow the inverse model to follow them. To autoselect the best generated plans, we use a weighted combination of a transition classifier and an object detector. The classifier is trained using image pairs that are within 1-step apart as positive labels, and random image pairs as negative labels.
In general, generative models like GAN and VAE suffer from consistently generating the same number of objects in the scene \cite{zhao2018bias}. 
Similarly, we find that CIGAN models sometimes mistakenly produce plans with extraneous blocks. Thus, we use the Mask-RCNN as an object detector \cite{matterport_maskrcnn_2017, IEEEexample:maskrcnn} to restrict this behavior. The Mask-RCNN is trained to detect randomly generated shapes which are agnostic to this domain. The object detector gives a binary value which is 1 if the object detector finds only one object (which is realistic), and 0 otherwise. 

We give an image pair by a score $s$ such that $s(o, o_{next}) = (\beta + c(o, o_{next}))^2 +  \alpha f(o,o_{next})$ where $c$ is the classifier score and $f$ is the the object detector score. For our purposes, we found the most promising results when $\alpha = 3.0, \beta = 1.0$. 

\subsection{A* Search Details}\label{app:astar}
In the rope domain, we performed A* search in the latent space using the Euclidean distance as the cost function and the heuristic. The graph search is done by expanding the node in the computation stac with the minimum cost-so-far plus $\lambda$ times cost-to-go, evaluated using the heuristic. Note that $\lambda$ is the relaxation term that we set at 1.4. The relaxation term controls the trade off between the close-to-optimal path and the computation time. The larger the lambda the greedier the search becomes. 
To expand a node, we find neighbors by doing Monte Carlo sampling of the transition function to find 50 neighbors. Finally, to check whether we have reached the goal, we threshold any distance below 1.5 in the latent space as sufficient.

\subsection{Network Architectures}
In Tables \ref{tab:inv_arch} and \ref{tab:cigan_arch}, we outline the architectures used for the CIGAN and inverse control models.

\begin{table}[h!]
\caption{Architecture of CIGAN used for rope domain}
\centering
\begin{tabular}{ |p{4.5cm}|p{4.5cm}| }
 \hline
 Discriminator D & Generator G\\
 \hline
 Input 2 64 x 64 grayscale images & Input a latent vector in $\mathbb{R}^{10}$\\  
 \hline
 4 x 4 conv. 128 lReLU, stride 2, batchnorm & 4 x 4 upconv. 1024 lReLU, stride 2, batchnorm \\
 \hline
 4 x 4 conv. 256 lReLU, stride 2, batchnorm & 4 x 4 upconv. 512 lReLU, stride 2, batchnorm \\
 \hline
 4 x 4 conv. 512 lReLU, stride 2, batchnorm & 4 x 4 upconv. 256 lReLU, stride 2, batchnorm \\
 \hline
  4 x 4 conv. 1024 lReLU, stride 2, batchnorm & 4 x 4 upconv. 128 lReLU, stride 2, batchnorm \\
  \hline
  4 x 4 conv. 1 & 4 x 4 upconv. 2 Tanh \\ 
  \hline
\end{tabular}
\label{tab:cigan_arch}
\end{table}

\begin{table}[h!]
\caption{Architecture of Inverse Model}
\centering
\begin{tabular}{|c| }
 \hline
 Input 1 64 x 64 image of pixelwise difference\\ 
 \hline
 4 x 4 conv. 64 lReLU, stride 2, dropout (.5) \\
 \hline
 4 x 4 conv. 128 lReLU, stride 2, dropout (.5) \\
 \hline
 4 x 4 conv. 256 lReLU, stride 2, dropout (.5) \\
 \hline
  4 x 4 conv. 512 lReLU, stride 2, dropout (.5)  \\
  \hline
  4 x 4 conv. 40, Sigmoid \\
  \hline
 40 dimensional linear, Tanh \vline \hspace{90pt} \\
 \hline 
 2 dimensional Linear, ReLu \vline \hspace{90pt} \\
 \hline 
 \hspace{92pt} \vline 42 dimensional linear, Tanh \\
 \hline 
 \hspace{92pt} \vline 2 dimensional linear, ReLu \\
 \hline 
\end{tabular}\\
\label{tab:inv_arch}
\end{table}

\subsection{Datasets}
Block wall dataset available here:
\\
\href{https://drive.google.com/drive/folders/16L3Bir66Y31O0khBKAntsxoxq_94U1T_?usp=sharing}{https://drive.google.com/drive/folders/16L3Bir66Y31O}
\\
\href{https://drive.google.com/drive/folders/16L3Bir66Y31O0khBKAntsxoxq_94U1T_?usp=sharing}{0khBKAntsxoxq\_94U1T\_?usp=sharing}.

Rope datasets available here:
\\
\href{https://drive.google.com/drive/folders/1zUQWAzxzN5-WvMc3u9C9GwWCrFb6UjVK?usp=sharing}{https://drive.google.com/drive/folders/1zUQWAzxzN5-WvMc3u9C9GwWCrFb6UjVK?usp=sharing}.

\subsection{Additional Block-Wall Results}

\begin{figure*}
  \centering
  \parbox{9.9in}{\includegraphics[scale=.6]{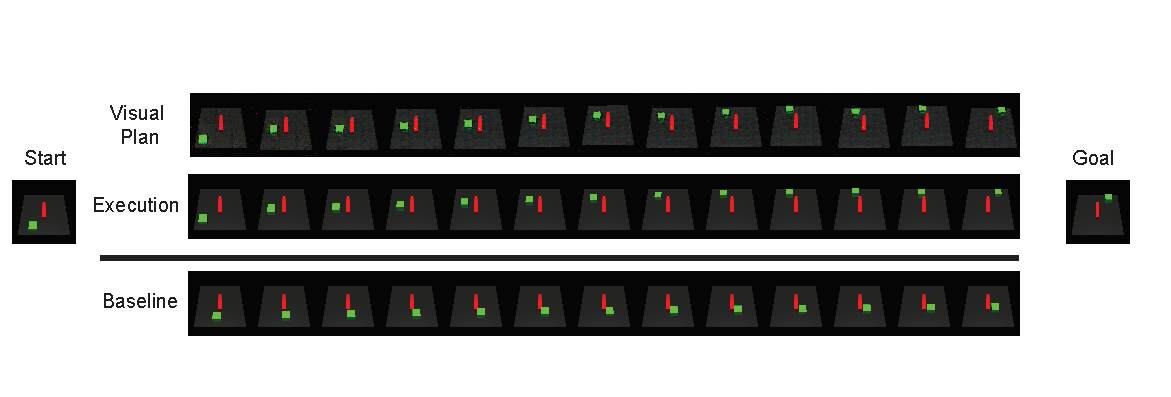}
}
  \caption{Another example of a start/goal test image where planning via VPA is essential in reaching our desired state. Our baseline, on bottom, is unable to navigate past the bottom right corner, while the plan generated by CIGAN finds a path around the obstacle rather than trying to go through.}
  \label{blockwall5}
 \end{figure*}
 \begin{figure*}
 \centering
  \parbox{9.9in}{\includegraphics[scale=.6]{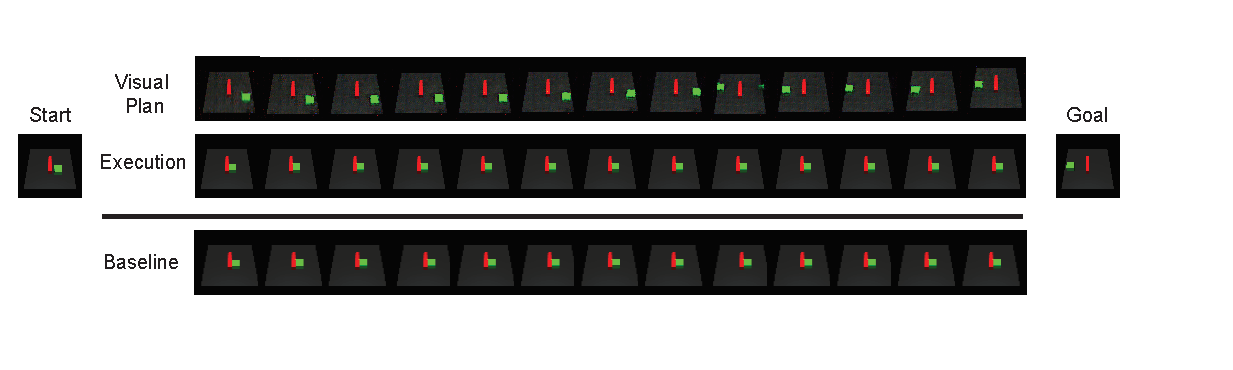}
}
  \caption{Example of failure case. While CIGAN succeeded 90\% of the time in our autoselected evaluation, there were a few cases that our approach failed to find a viable plan, and thus we were unable to successfully reach the goal state during execution.}
  \label{blockwall16}
\end{figure*}
\end{appendix}

\end{document}